\documentclass{article}

\newcommand{\keywords}[1]{%
  \vspace{1em}%
  \noindent\textbf{Keywords: } #1%
}

\usepackage{geometry}
\usepackage{graphicx}%
\usepackage{multirow}%
\usepackage{amsmath,amssymb,amsfonts}%
\usepackage{amsthm}%
\usepackage{mathrsfs}%
\usepackage[title]{appendix}%
\usepackage{xcolor}%
\usepackage{textcomp}%
\usepackage{manyfoot}%
\usepackage{booktabs}%
\usepackage{algorithm}%
\usepackage{algorithmicx}%
\usepackage{algpseudocode}%
\usepackage{listings}%

\begin{document}
\title{Image-Intrinsic Priors for Integrated Circuit Defect Detection and Novel Class Discovery via Self-Supervised Learning}
\author{Botong.Zhao,Xubin.Wang,Shujing.Lyu,Yue.Lu}
\date{}

\maketitle

\abstract{Integrated circuit manufacturing is highly complex, comprising hundreds of process steps. Defects can arise at any stage, causing yield loss and ultimately degrading product reliability. Supervised methods require extensive human annotation and struggle with emergent categories. We propose IC-DefectNCD, a support-set-free framework that leverages Image-Intrinsic Priors in IC SEM images for defect detection and novel class discovery. First, Self-Normal Information Guided IC Defect Detection is introduced, which aggregates representative normal features through a learnable normal information extractor and uses reconstruction residuals to coarsely localize defect regions. To handle saliency variations across defects, an adaptive binarization strategy is adopted to produce stable subimages focused on defective areas. Finally, Self-Defect Information Guided IC Defect Classification is developed, which incorporates a soft-mask guided attention to inject spatial defect priors into a teacher-student model, enhancing sensitivity to defective regions and suppressing background interference. This enhances sensitivity to defective regions, suppresses background interference, and enables recognition and classification of unseen defects.
The proposed approach is validated on a real-world dataset spanning three key fabrication stages and covering 15 defect types. Experiments demonstrate robust performance on both defect detection and unseen defect classification.}


\section{Introduction}
\label{sec:introduction}
With the rapid advancement of semiconductor manufacturing, Integrated circuit(IC) technology nodes continue to shrink and circuit density continue to rise, making fabrication increasingly complex.
During production, even minute anomalies at any process step can manifest as surface defect.
As shown in Fig. 1, wafer images captured by scanning electron microscopy (SEM) typically exhibit complex backgrounds, extremely low defect rates, and marked diversity in defect morphology.
These defects indicate process excursions and equipment health, and are tightly linked to final device performance and reliability.
Efficient wafer defect detection and management are therefore essential for improving manufacturing yield.

\begin{figure}[h]
        \centering
	\includegraphics[width=0.65\textwidth]{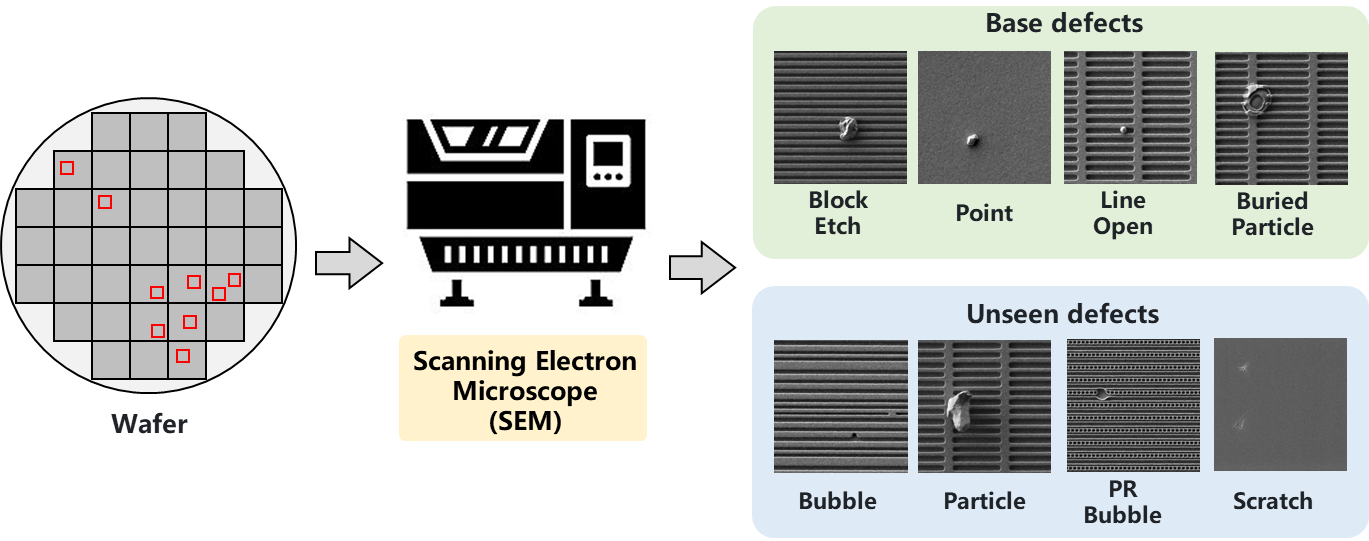}
	\caption{Wafer site sampling and SEM imaging pipeline, with layout changes in IC manufacturing continuously introducing unseen defects.}
	\label{Fig:1}       
\end{figure}

Therefore, semiconductor fabs typically employ a large team of yield engineers dedicated to defect inspection and analysis. Engineers use SEM images to closely examine wafers, then quantify, analyze, and classify observed defects by type, size, and appearance. Engineers infer root causes and coordinate with the process-module owners, making SEM-based defect analysis integral to every stage of IC manufacturing.

Although AI methods improve detection efficiency \cite{hu2020lightweight, zeng2021reference, 11180178, 10410042, 10454019}, most approaches rely on the unrealistic assumption that defects and backgrounds at inference time must have appeared in the training or support set. This hinders deployment in real production environments. As shown in Fig. 2, supervised models \cite{qiao2024deepsem, mei2025novel, zhao2024integrated} classify known defects accurately but fail to generalize to unknown categories. Unsupervised methods can discover anomalies, yet their classification performance is weak and the support set cannot cover complex and evolving process backgrounds. Fundamentally, these approaches adopt a static design that cannot adapt to continually emerging backgrounds and defects driven by new processes, materials, and layouts in production. In addition, the uniqueness, scarcity, and confidentiality of SEM images make large-scale annotation difficult. Therefore, fast adaptation for detecting and classifying unseen defect categories is a core requirement for practical industrial deployment.

\begin{figure}[h]
        \centering
	\includegraphics[width=0.6\textwidth]{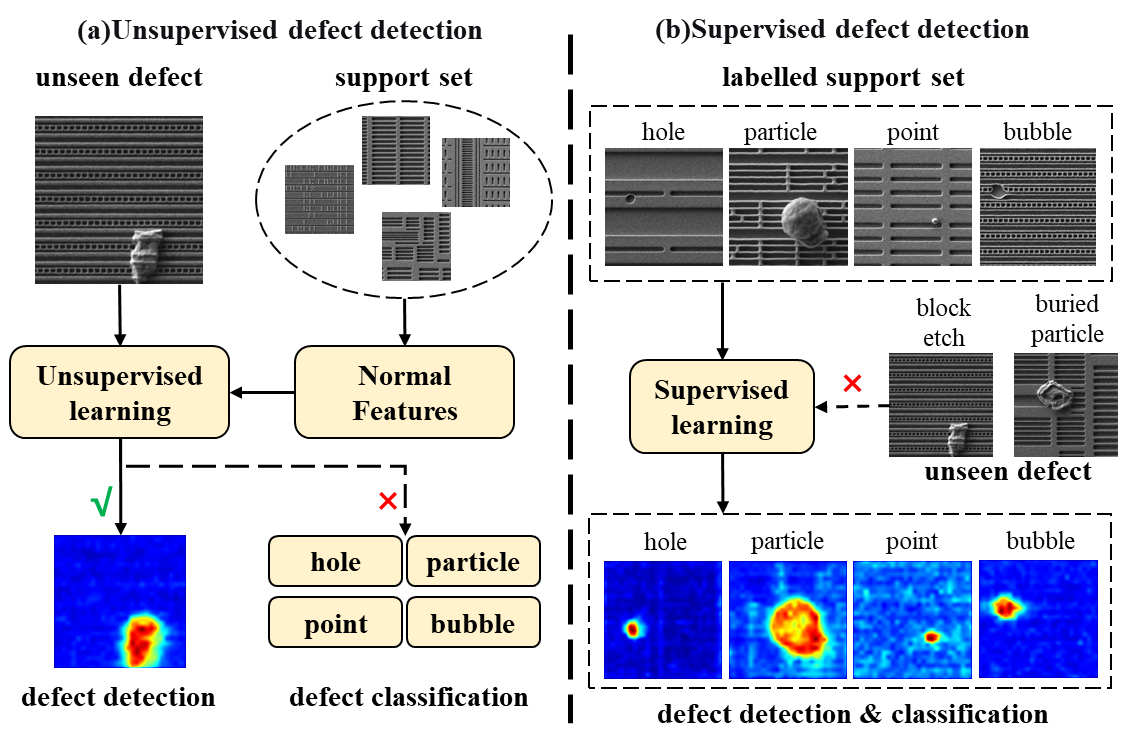}
	\caption{ Comparison of unsupervised and supervised defect detection. (a) Unsupervised methods use normal features to localize anomalies but cannot categorize unseen defects. (b) Supervised methods classify known defects from labeled data but fail on unseen categories and demand extensive annotation.}
	\label{Fig:1}       
\end{figure}

Building on these observations, we identify two key differences between IC scenarios and conventional industrial defect detection. First, IC SEM images have highly complex backgrounds, and the task must both localize defects and determine their categories. Second, within a single IC SEM image, normal regions are strongly correlated in structure, and defect regions show a stable contrast to the image’s normal patterns. These characteristics mean that localization alone or cross-image priors are insufficient, and leveraging normal information within each image is essential for detection and classification.

We present IC-DefectNCD, a unified framework that leverages single-image priors for IC defect detection and novel class discovery. First, a learnable Normal-Information Extractor (NI-Extractor) aggregates single-image normal informations, reconstructs normal content, and localizes defects via reconstruction residuals. Then, Self-Saliency-Driven Adaptive Binarization produces stable, defect-centered crops. Finally, Soft-Mask–Guided Attention (SMG-Attention) injects spatial defect priors into the teacher–student module to focus representations on defects and suppress background. Semi-supervised k-means estimates the number of unseen defect classes and configures the classification head. The system delivers robust detection and accurate classification of both base and unseen classes.

The primary contributions of this research are outlined as follows:

1. We present IC-DefectNCD, a unified framework for IC defect detection and novel class discovery that exploits single-image internal cues in SEM imagery and removes any support-set dependency at inference.

2. A self-saliency-driven adaptive binarization with center cropping leverages the relative saliency between defective and normal regions to adaptively segment defects in IC SEM images.

3. SMG-Attention injects a soft-mask spatial prior into a teacher-student self-supervised framework, strengthening focus on defective regions and enabling reliable identification and classification of unseen defects.

4. Validated on real data spanning BEOL, DEP, and DPR with 15 defect types, the approach delivers strong detection and classification performance with a simple, support-set-free inference pipeline, demonstrating robust generalization.

\section{Related work}
\subsection{IC defect detection}
Due to early technological limitations, wafer-surface defect inspection primarily relied on manual visual checks. Faced with increasingly complex IC designs, this approach demands substantial manpower and suffers from low efficiency, limited accuracy, and frequent misses and false alarms. With technological progress, traditional computer vision techniques have been applied to defect detection, such as edge detection \cite{zhao2010multi}, morphological operations \cite{hu2016detection}, and texture feature extraction \cite{zuo2012fabric}. In parallel, machine learning–based methods have also been explored, most commonly feature-driven schemes using descriptors like gray-level co-occurrence matrices and wavelet transforms. However, these approaches depend heavily on hand-crafted rules and expert knowledge, making them ill-suited to the diverse, subtle, and rapidly evolving defect types in IC manufacturing \cite{hwang2007model, li2006improving}.

In recent years, driven by deep learning, defect detection has shifted toward more intelligent paradigms with several mainstream directions. Reconstruction-based methods learn the distribution of normal samples and localize anomalies via reconstruction error, for example transparency-guided residual inpainting in TransFusion \cite{fuvcka2024transfusion}, multi-scale latent restoration and contrast in DiAD \cite{he2024diffusion}, and adaptive-step denoising for varying saliency in GLAD \cite{yao2024glad}. Memory-bank–based methods extract features with pretrained models and detect anomalies by comparing against stored normal features, such as multi-scale local mutual scoring in MuSc \cite{li2024musc}, patch-level noise discrimination with denoising before building the core memory in SoftPatch \cite{jiang2022softpatch}, and DINOv2-based patch similarity for few-shot detection in AnomalyDINO \cite{damm2025anomalydino}. In addition, with the rise of vision–language models, multimodal approaches align textual prompts with localized visual features for anomaly localization and recognition, including PromptAD \cite{li2024promptad}, AA-CLIP \cite{ma2025aa}, AnomalyCLIP \cite{zhou2023anomalyclip}, FiLo \cite{gu2024filo}, and MultiADS \cite{sadikaj2025multiads}.

These general methods face inherent challenges on IC SEM images. Unlike conventional industrial products with uniform appearance, IC SEM imagery exhibits highly structured and strongly repetitive textures, and defects often manifest as subtle, localized anomalies embedded in the background. This characteristic misaligns with the global semantic cues that generic vision models rely on, leading to substantial challenges in stability and cross-process transferability for IC defect detection.

To address these domain-specific challenges, several studies have explored IC-tailored solutions. Lu leverages Masked Autoencoders (MAE) for reconstruction-based segmentation of SEM defects \cite{lu2023masked}. Chen proposes a CNN–Transformer hybrid framework for IC defect segmentation \cite{mei2025novel,qiao2024deepsem}. Jiang investigates trainable multimodal large models to align IC defect semantics \cite{jiang2024fabgpt}. Despite progress, these approaches still rely on support sets of normal samples and substantial manual annotation, and their generalization across layouts and process stages remains uncertain. Sadikaj et al.\ use differences in text features across defect categories to label masks and thereby classify defects \cite{sadikaj2025multiads}. Zhao introduces a multi-scale receptive-field mechanism to classify defects of varying sizes\cite{zhao2024integrated}.

However, a fundamental limitation remains. Most methods assume a closed world in which defects and backgrounds observed at inference are predefined in the training phase or the support set. This assumption conflicts with the dynamic reality of IC production, where evolving processes and layouts continually introduce unseen backgrounds and defect categories. Consequently, unseen defect classes remain a challenge for existing methods, constraining their practical deployment in IC manufacturing.

We take a different direction aligned with IC manufacturing. Instead of relying on support sets and extensive labels, the approach mines the internal information of each test image to improve generalization in defect detection. Concretely, normal information within a single image enables highly generalizable detection, while self-supervised learning combined with semi-supervised k-means discovers and classifies unknown defect categories. This design better matches fast iteration and open-world uncertainty in IC production.

\subsection{Binarization}
Binarization is a key preprocessing step in many vision tasks and has been widely used in edge detection \cite{lee2023uniformaly,khosla2020supervised}, image segmentation \cite{ji2019invariant,hsu2018learning,yang2023bootstrap}, document image processing \cite{han2019learning,chen2021exploring,salehani2020msdb}, medical image analysis \cite{zhou2024exploiting,ntirogiannis2013performance}, and object detection \cite{ahn2018learning,fini2021unified,vaze2022generalized}. A classical approach is Otsu’s method \cite{otsu1979threshold}, which adaptively selects a global threshold for separating foreground and background by maximizing inter-class variance. Recently, learning-based methods such as DiffuMask \cite{wu2023diffumask} leverage cross-attention maps together with adaptive thresholds to produce binary masks. Although general-purpose segmentation models such as SAM and SAM 2 show strong performance, their substantial computational cost, bias toward natural images, and limited adaptation under scarce IC SEM data constrain their practicality for our task \cite{ravi2024sam}. More importantly, IC defects are typically subtle local anomalies embedded within normal structures, which provide weak semantic cues, making it difficult for general models to reliably identify them. Existing binarization pipelines often produce scattered false-positive regions when applied to anomaly segmentation, which severely interferes with downstream region-level anomaly categorization and overlooks the pronounced saliency differences across defect types. To tackle these issues, we propose a self-saliency driven adaptive binarization that concentrates on extracting the principal connected anomalous regions, thereby suppressing spurious detections and improving the robustness of subsequent self-supervised classification.

\subsection{Novel Class Discovery}
Novel Class Discovery(NCD) aims to leverage knowledge learned from labeled classes to identify and cluster new classes in unlabeled data. Early frameworks were often formulated as deep transfer clustering \cite{han2019learning}. Most early methods followed a two-stage pipeline \cite{hsu2018learning,hsu2019multiclass,han2019learning}: first learn a representation prior on base classes, then transfer it to the novel set for clustering. More recently, one-stage methods \cite{autonovel_pami2021,openmix_cvpr2021,vaze_nips2023} jointly handle base and unseen data, reducing bias toward base classes and yielding more generalizable representations. However, these approaches typically assume that the number of unknown classes is known a priori and that the unlabeled set contains no samples from seen classes. This assumption conflicts with the realities of IC manufacturing: defect categories are dynamic and often include previously unseen types, and newly emerging and legacy defects frequently co-occur. Moreover, IC defects typically occupy only a small portion of an image, leaving their features easily overwhelmed by the extensive normal background. To overcome these limitations, we first introduce SMG-Attention to sharpen the model’s focus on defect regions while preserving contextual relations between defects and normal areas. Building on this, we propose a semi-supervised k-means inference strategy that automatically estimates the number of unknown classes, and we construct a self-supervised classifier enhanced with IC defect priors to effectively discover and categorize unseen defects.

\section{Proposed method}

\begin{figure*}[htb]
        \centering
	\includegraphics[width=0.9\textwidth]{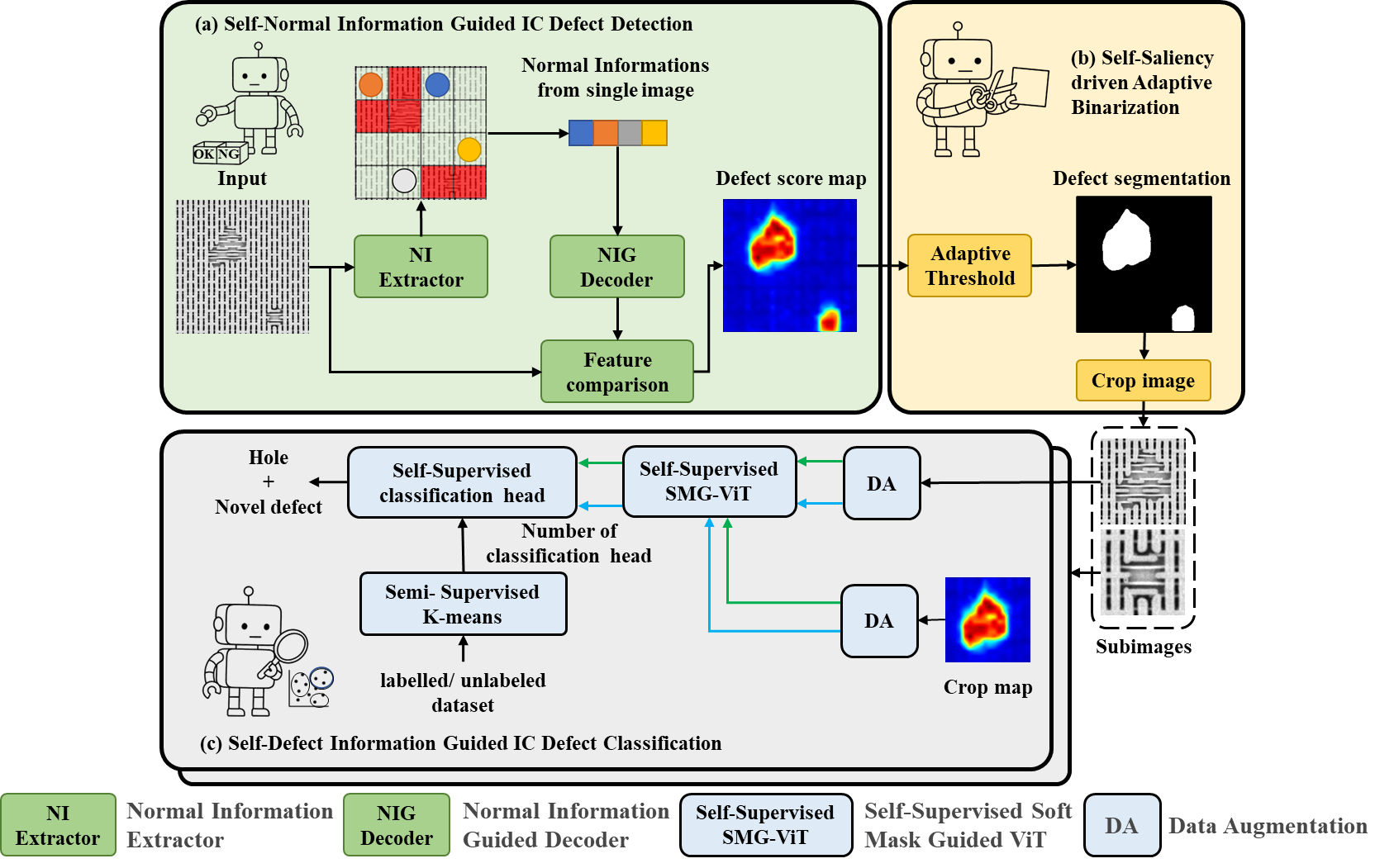}
	\caption{(a) \textbf{Self-Normal Information Guided IC Defect Detection:} NI-Extractor gathers normal informations and NIG-Decoder reconstructs patch tokens by these normal informations, and feature comparison yields a defect score map.(b) \textbf{Self-Saliency-driven Adaptive Binarization:} adaptive thresholding and center cropping produce a compact soft mask.(c) \textbf{Self-Defect Information Guided IC Defect Classification:} SMG-ViT uses the soft mask within teacher–student self-supervised learning, and semi-supervised k-means estimates the number of unknown classes and sets the number of classification heads for unseen defect classification.}
	\label{Fig:1}       
\end{figure*}

IC-DefectNCD focuses on mining single-image internal cues for robust defect detection and classification in wafer production. As shown in Fig. 3, the Self-Normal Information Guided IC Defect Detection stage reconstructs the normal component of the image and produces a residual-based defect score map. The Self-Saliency-driven Adaptive Binarization module then converts the score map into a compact soft mask using adaptive thresholding and center cropping. Finally, the Self-Defect Information Guided IC Defect Classification stage adopts SMG-ViT, which augments transformer self-attention with SMG-Attention to learn defect-centric representations in a teacher–student self-supervised framework. Semi-supervised k-means estimates the number of unknown classes in the unlabeled set and configures the classification head, enabling recognition and classification of unseen defects.

\subsection{Problem Formulation and Notation}
In IC defect detection practical applications, the unlabeled set $D_u$ usually contains both base(seen) and unseen classes, with the latent label space $Y_u = C_{\text{base}} \cup C_{\text{u}}$, where the cardinality of unseen classes $K_{\text{u}} = |C_{\text{u}}|$ is unknown a priori.
Our goal is to distinguish the base and unseen classes in $D_u$ and estimate the total number of classes, thereby providing reliable priors for subsequent learning of the unseen classes.
Following the standard NCD setting, we adopt a small set of labeled abnormal images
$D_l=\{(I_i^l, y_i^l, M_i^l)\mid i\in[1,N_l]\}$,
which contains $C_l$ available (“Base”) classes.
Here, $y_i^l \in \{0,1\}^{1\times(C_l+C_u)}$ is the one-hot category label,
and $M_i^l$ is the ground-truth anomaly mask of image $I_i^l$ in $D_l$.
The labeled set $D_l$ provides industrial priors that guide grouping within $D_u$ and the discovery of unseen defect classes.

\subsection{Self-Normal Information Guided IC Defect Detection}
Many approaches extract or learn “normal” features from a support set and then compare them against a test image. However, in IC manufacturing, these methods encounter fundamental limitations. IC SEM images exhibit highly structured backgrounds, and attention’s global semantics and positional encodings can distort local evidence, making support-derived “normal” features difficult to align with local test features. We observe that intra-image features are more strongly correlated than features matched from an external support set. Moreover, constructing a support set that spans diverse layouts, processes, and revisions is impractical in real production.

\begin{figure*}[h]
        \centering
	\includegraphics[width=0.9\textwidth]{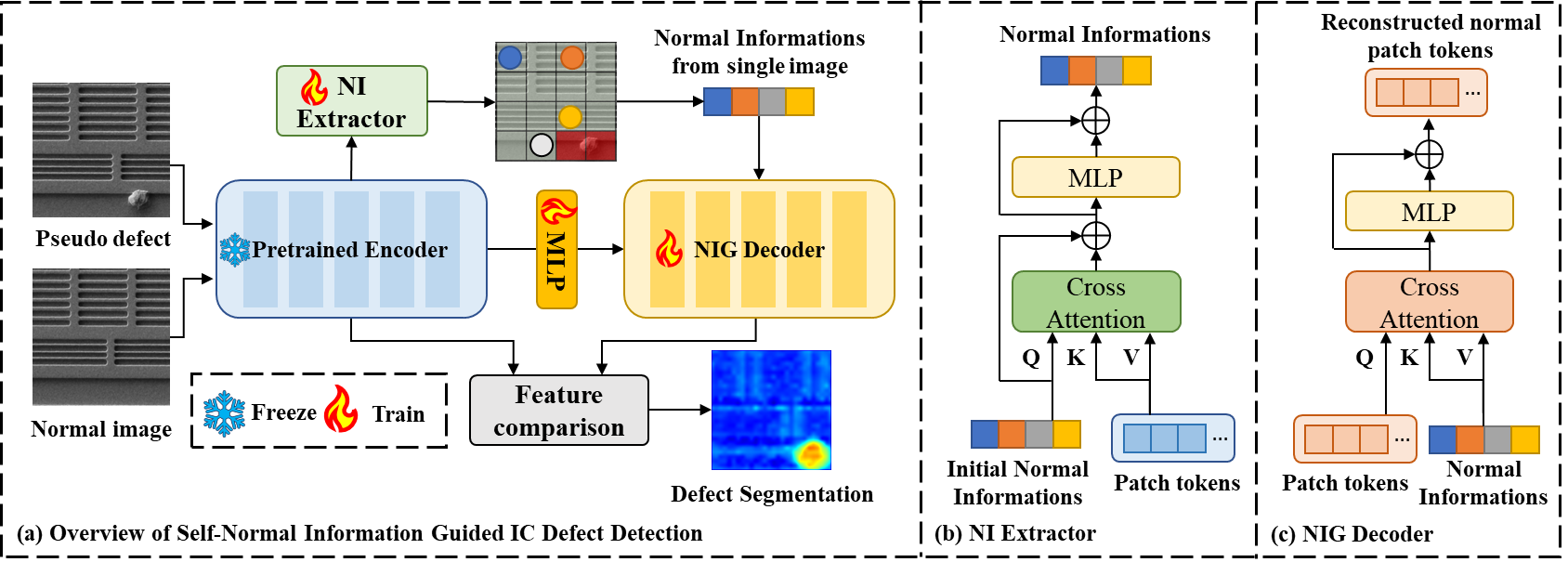}
	\caption{(a) \textbf{Overview of the proposed Self-Normal Information Guided IC Defect Detection framework.} The framework enhances detection robustness by mining normal cues directly from a single image without relying on external support sets. It reconstructs normal patterns guided by these cues and isolates abnormal regions through residual analysis. (b) Detailed architecture of each layer in the NI Extractor. (c) Detailed architecture of each layer in the NIG Decoder.}
	\label{Fig:1}       
\end{figure*}

To address this challenge, we propose Self-Normal Information Guided IC Defect Detection. As illustrated in Fig. 4(a), the framework has two stages: (i) extraction of representative normal features and (ii) normal-information-guided reconstruction. First, a learnable Normal-Information (NI) extractor aggregates features that characterize the image’s intrinsic normal patterns. Next, a Normal-Information-Guided(NIG) Decoder uses these features to reconstruct local features. Finally, residuals between reconstructed and original features provide a defect score, which coarsely localizes and segments candidate regions.

The learnable NI-Extractor mines single-image normal informations directly from the input, eliminating any reliance on an external support set. We synthesize pseudo defects and corresponding binary masks by randomly masking features extracted from normal samples on the training images, yielding paired normal and pseudo-abnormal samples that mimic production cases. Next, a pretrained backbone extracts multi layer features$\{f_\ell\}_{\ell=1}^{L}$, and each feature map satisfies $\text{size}(f_\ell) = N \times C$. Here, $N$ denotes the number of patch tokens, and $C$ denotes the feature dimension.

As shown in Fig. 4(b), we aggregate features from multiple layers to obtain $F$, a multi-scale representation that serves as the keys and values of the normal-information extractor. Next, we apply cross-attention by taking $M$ initialized learnable normal tokens $F_{\text{normal}}^{\text{init}}$ together with $F$ as input, and obtain $F_{\text{normal}} \in R^{M \times C}$, which encodes the information of normal regions. The computation is as follows.

\begin{gather*}
\mathbf{F} = \sum_{\ell=1}^{L}\mathbf{f}^{\,\ell},\\
\mathbf{Q} = \mathbf{F}_{\text{normal}}^{\text{init}}\mathbf{W}_{q},\quad
\mathbf{k} = \mathbf{F}\mathbf{W}_{k},\quad
\mathbf{v} = \mathbf{F}\mathbf{W}_{v},\\
\mathbf{F}'_{\text{normal}} = \operatorname{Attention}(\mathbf{Q},\mathbf{k},\mathbf{v}) + \mathbf{F}_{\text{normal}},\\
\mathbf{F}_{\text{normal}} = \operatorname{MLP}(\mathbf{F}'_{\text{normal}}) + \mathbf{F}'_{\text{normal}}.
\end{gather*}

$W_q$, $W_k$, and $W_v \in R^{C \times C}$ denote the learnable parameters in the cross-attention module, and $MLP$ denotes a multilayer perceptron.

To ensure that the extracted normal information truly originates from normal regions, we design the following loss function. First, a downsampled mask is employed to separate normal regions from pseudo defect areas. Next, we compute the cosine distance between the extracted normal information and the features from both normal and abnormal areas. The objective is to minimize the distance to features of normal areas while maximizing the distance to those of abnormal areas. This encourages the extracted normal information to remain strongly correlated with the true normal regions. The loss function is defined as follows.

\[
\begin{aligned}
d_i^{+} &= \min_{m\in\{1,\ldots,M\}}
\operatorname{distance\_cos}\!\big(\mathbf{F}(i),\,\mathbf{F}_{\text{normal}}^{\,m}\big),\\
d_j^{-} &= \min_{m\in\{1,\ldots,M\}}
\operatorname{distance\_cos}\!\big(\mathbf{F}(j),\,\mathbf{F}_{\text{normal}}^{\,m}\big),\\
\mathcal{L}_{normal} &= \frac{1}{|\Omega_{\mathrm{n}}|}\sum_{i\in\Omega_{\mathrm{n}}} d_i^{+}
+\frac{1}{|\Omega_{\mathrm{a}}|}\sum_{j\in\Omega_{\mathrm{a}}} (1-d_j^{-}),
\qquad \alpha\ge 0,
\end{aligned}
\]

where $\operatorname{distance\_cos}(\cdot,\cdot)$ denotes the cosine similarity between the patch tokens extracted by the pretrained model and the normal-information tokens $F_{\text{normal}}$. Here, $d_i^{+}$ denotes the distance to the nearest normal patch token for $i\in\Omega_{n}$, and $d_j^{-}$ denotes the distance to the nearest abnormal patch token for $j\in\Omega_{a}$.

It is important to note that the extracted normal information is not a single local feature but a cross-regional representation capturing the overall normal pattern of the image. Therefore, directly computing patch-level feature distances for anomaly scoring using this information is unreliable. Instead, we employ the extracted normal information to guide the reconstruction of IC features. Defect regions are subsequently segmented by computing residuals between the reconstructed and original features. This process effectively suppresses global bias while amplifying fine-grained deviations from the normal structure.

As shown in Fig. 4(c), ${F}_{\text{normal}}$ is injected into the decoder to guide feature reconstruction. Because ${F}_{\text{normal}}$ represents only normal region features, we use it as the keys and values. During training, the decoder focuses on reconstructing features from normal regions, which effectively suppresses reconstruction for anomalous queries.

\begin{gather*}
\mathbf{Q}_{\ell} = \mathbf{f}_{\text{decoder}}^{\,\ell-1}\mathbf{W}^{Q}_{\ell},\quad
\mathbf{K}_{\ell} = \mathbf{F}_{\text{normal}}\mathbf{W}^{K}_{\ell},\quad
\mathbf{V}_{\ell} = \mathbf{F}_{\text{normal}}\mathbf{W}^{V}_{\ell},\\
\mathbf{A}_{\ell} = \operatorname{ReLU}\!\big(\mathbf{Q}_{\ell}\mathbf{K}_{\ell}^{\top}\big),
\mathbf{f}_{\text{decoder}}^{\,\ell-1'} = \mathbf{A}_{\ell}\mathbf{V}_{\ell},\quad,\\
\mathbf{f}_{\text{decoder}}^{\,\ell} = \operatorname{MLP}\!\big(\mathbf{f}_{\text{decoder}}^{\,\ell-1'}\big)
+ \mathbf{f}_{\text{decoder}}^{\,\ell-1'} .
\end{gather*}

Here, $\mathbf{f}_{\text{decoder}}^{\,\ell}\in\mathbb{R}^{N\times C}$ denotes the output of the $\ell$-th decoder layer, and $\mathbf{W}^{Q}_{\ell}, \mathbf{W}^{K}_{\ell}, \mathbf{W}^{V}_{\ell}\in\mathbb{R}^{C\times C}$ are the learnable parameters of the $\ell$-th decoder layer. 
To make the reconstructed normal features closely match the encoder’s original normal features while emphasizing hard regions during backpropagation and suppressing gradient interference from easy regions, we design the loss to directly modulate feature gradients.

\[
\begin{aligned}
w^{\ell}(h,w) &= \left[\frac{M^{\ell}(h,w)}{u\!\left(M^{\ell}\right)}\right]^{\gamma},\\[4pt]
\mathcal{L}_{reconstruction} &= \frac{1}{L}\sum_{\ell=1}^{L}
\operatorname{distance\_cos}\!\Big(\operatorname{vec}(\mathbf{f}^{\,\ell}),\,
\operatorname{vec}(\hat{\mathbf{f}}_{D}^{\,\ell})\Big),\\[4pt]
\hat{\mathbf{f}}_{D}^{\,\ell}(h,w) &= cg\!\big(\mathbf{f}_{D}^{\,\ell}(h,w)\big)\; w^{\ell}(h,w).
\end{aligned}
\]

where $u(M^{\ell})$ represents the average regional cosine distance within a batch, $\gamma \ge 0$ denotes the temperature hyper-parameter, $cg(\cdot)\, w^{\ell}(h,w)$ denotes a gradient adjustment based on the dynamic weight $w^{\ell}(h,w)$, and $\operatorname{vec}(\cdot)$ denotes the flattening operation. The overall training loss of our INP-Former can be expressed as

\[
\mathcal{L}_{\text{total}} \;=\; \mathcal{L}_{reconstruction}\;+\; \lambda\,\mathcal{L}_{normal}.
\]

\subsection{Self-Saliency-driven Adaptive Binarization}
The reconstruction residuals yield a defect \emph{score map} that serves as a coarse prior for localization.
As shown in Fig. 5, binarization across different thresholds reveals that defects vary in visual saliency, and the \emph{number of detected defect regions fluctuates noticeably} as the threshold decreases.
This instability arises because defect regions typically receive higher defect scores than normal regions, while normal regions, although more consistent, still exhibit inherent fluctuations.
These observations motivate an image-specific threshold rather than a fixed one.

\begin{figure}[h]
        \centering
	\includegraphics[width=0.8\textwidth]{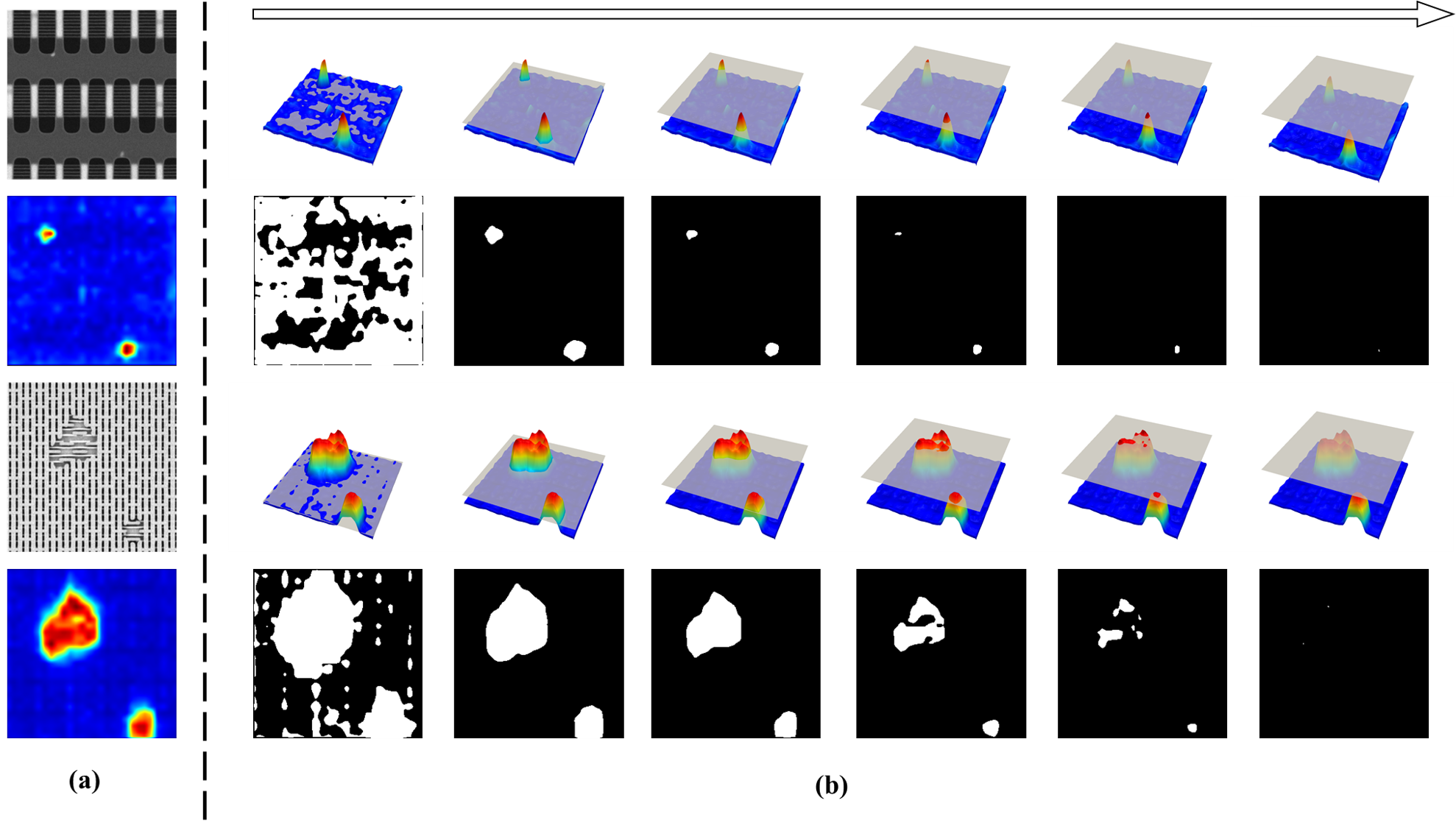}
	\caption{Binarization results under different thresholds.}
	\label{Fig:1}       
\end{figure}

Let $R\in\mathbb{R}^{H\times W}$ denote the residual map.
We compute a \emph{self-saliency} map $S\in[0,1]$ by percentile normalization

\[
S=\mathrm{clip}\!\left(\frac{R-\mathrm{perc}_{p_1}(R)}{\mathrm{perc}_{p_2}(R)-\mathrm{perc}_{p_1}(R)},\,0,\,1\right),
\]

where $\mathrm{perc}_{p}(R)$ is the $p$-th percentile.
We then sweep a descending threshold set $\mathcal{T}=\{t_0>t_1>\dots>t_K\}$ uniformly sampled in $[0,1]$ (e.g., $K=64$).
For each $t\in\mathcal{T}$ we form a binary mask $B_t=\mathbf{1}[S\ge t]$ and count its \emph{connected components} $c(t)$.
Fig. 6 plots $c(t)$ versus the threshold and shows a characteristic pattern where a stable plateau transitions to rapid fluctuation as the threshold decreases.

\begin{figure}[h]
        \centering
	\includegraphics[width=0.65\textwidth]{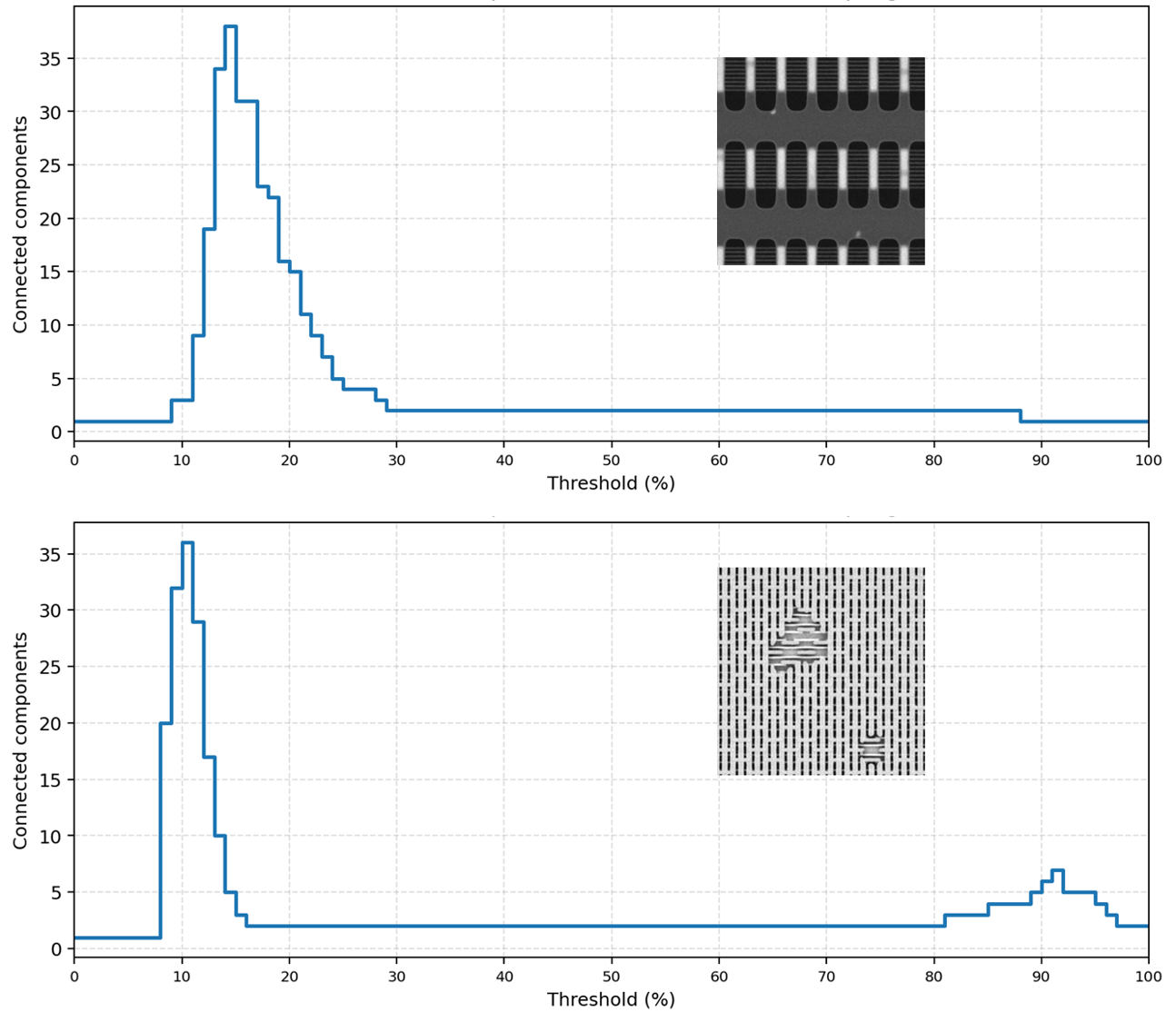}
	\caption{Number of defect regions after binarization with different thresholds.}
	\label{Fig:1}       
\end{figure}

We define a \emph{stable plateau} $\mathcal{P}\subseteq\mathcal{T}$ as a maximal consecutive index interval satisfying

\[
\lvert c(t_{k+1})-c(t_k)\rvert \le \varepsilon,
\]

with $\varepsilon\in\{0,1\}$.
Among all candidate plateaus, we select the longest one $\mathcal{P}^\star$.
If multiple plateaus tie, we choose the interval with the highest average inter-threshold $\mathrm{IoU}\big(B_{t_{k}},B_{t_{k+1}}\big)$ between consecutive masks.
The \emph{adaptive threshold} is the lowest value on the selected plateau,

\[
t^\star=\min \mathcal{P}^\star,
\]

which yields the most inclusive mask while remaining within the stable regime highlighted in Fig. 6.
The final mask is

\[
M=\mathrm{post}\!\big(B_{t^\star}\big),
\]

\subsection{Self-Defect Information Guided IC Defect Classification}
Unlike conventional classification, where models typically process complete and independent foreground objects, defects in integrated circuits take markedly different forms. Beyond a small number of independent particulate defects, most defects such as shorts and bubbles manifest as abnormal deformations of circuit structures, whereas opens and voids correspond to missing parts of specific structures. Pretrained ViT rely on global semantics learned from natural images and adapt poorly to the highly structured, low-saliency, and locally dominated defect patterns in IC-SEM. Under data-scarce production conditions, their transferability and robustness are often constrained.
Consequently, during classification the model must precisely focus on the defect region while capturing the transitions and contextual relationships between the defect and the surrounding normal circuitry.

\subsubsection{Self-supervised feature extractor guided by a soft mask}

\begin{figure*}[h]
        \centering
	\includegraphics[width=0.9\textwidth]{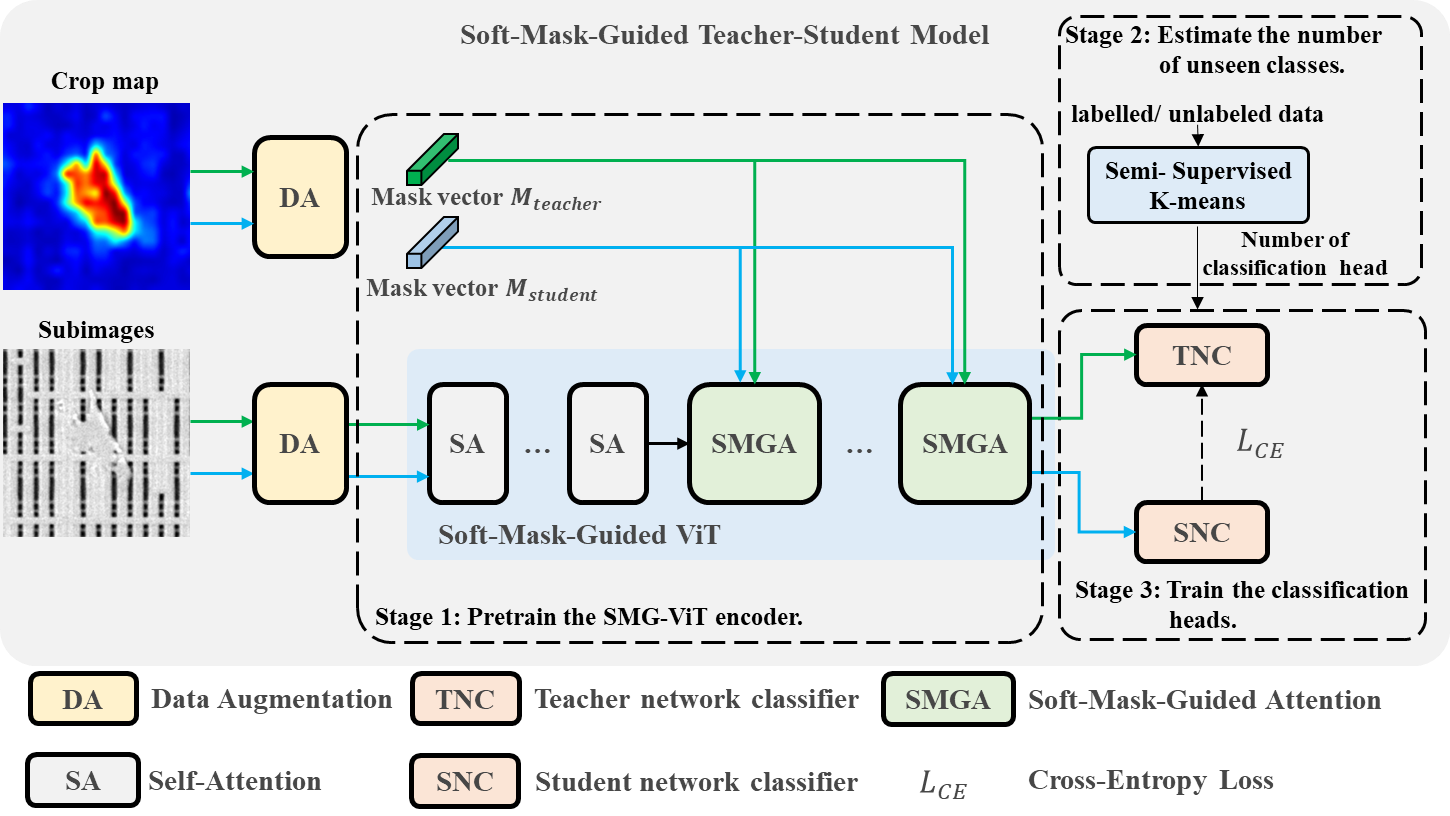}
	\caption{\textbf{Overview of the Soft-mask–guided teacher–student model.} \textbf{Stage 1} pretrains the SMG-ViT encoder with self-supervision to learn single-image priors and focus representations on defect regions. \textbf{Stage 2} estimates the number of unseen classes using semi-supervised k-means on mixed labeled and unlabeled features and configures the discovery head. \textbf{Stage 3} trains the classification heads within a teacher–student framework to classify both base and unseen defect classes.}
	\label{Fig:1}       
\end{figure*}

To address limited defect awareness and the difficulty of balancing local focus with global context, we introduce SMG-Attention after obtaining a defect-centered crop. This mechanism injects the defect heatmap as a spatial prior into the attention computation, guiding the model to attend to defects while preserving their linkage to normal regions. The computation of SMG-Attention is as follows.

\[
\hat{m}
=\Big[\,\max(M)\;;\;{vec}\!\big({AvgPool}_{p\times p}(M)\big)\Big]
\in {R}^{N+1},
\]
\[N=\frac{H}{p}\frac{W}{p},\]
\[
Attn=\operatorname{softmax}\!\big(Q_{l-1}K_{l-1}^{\top}+{Repeat}(\hat m)\big)\,V_{l-1}.
\]
Here $M$ denotes the defect segmentation heatmap of the cropped image. After downsampling and flattening, it is aligned with the patch tokens. The maximum heatmap value is inserted at the first position to provide a whole-image defect score aligned with the $[\mathrm{CLS}]$ token. Next, $\mathrm{Repeat}(\cdot)$ replicates the soft-mask vector across query rows, and the replicated vector is added as an additive bias in the attention computation. Notably, soft-mask--guided attention replaces only the last $j$ transformer layers to construct \textbf{SMG-ViT}. The remaining layers use DINOv2-pretrained ViT parameters and are frozen during fine-tuning.

Following DINO~\cite{oquab2023dinov2}, we apply distinct data augmentations to each defect-centered sub-image to obtain a pair $(x_{n,i}, x'_{n,i})$, which is then fed to a teacher--student model. This training strategy strengthens the sensitivity of \textbf{SMG-Attention} to anomalous regions. The training pipeline is illustrated in Fig.7.

As shown in Fig. 8, visualizing the last layer verifies the effectiveness of SMG-ViT, which drives the model to focus on defect regions.

\begin{figure}[h]
        \centering
	\includegraphics[width=0.65\textwidth]{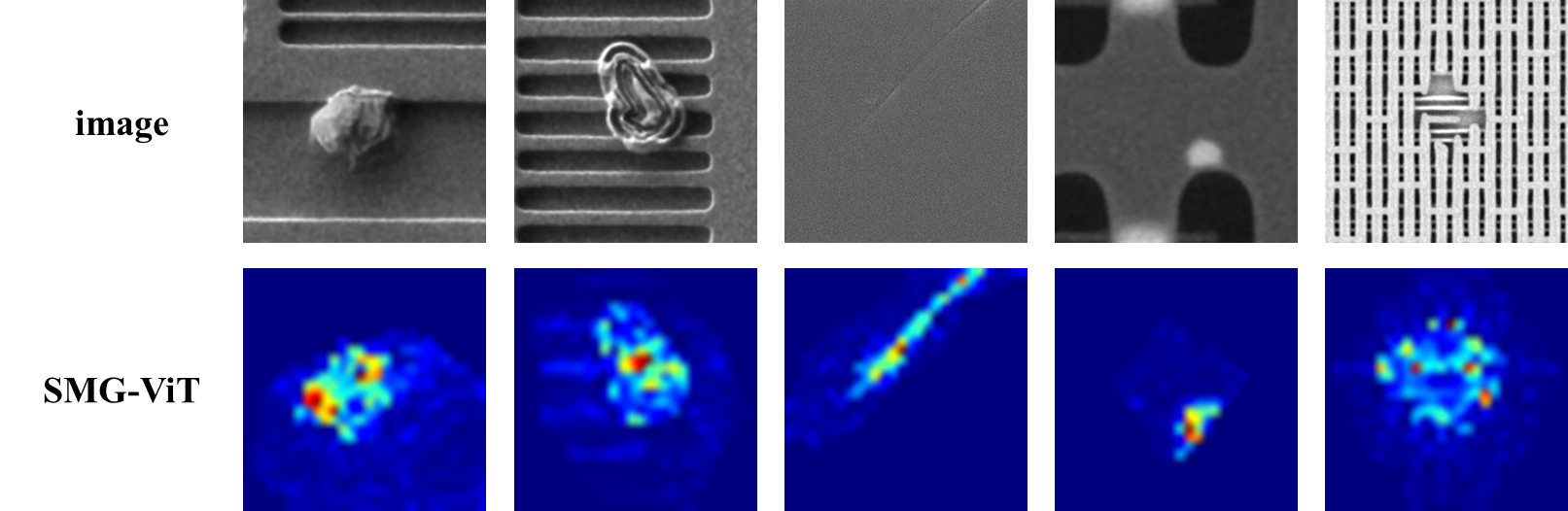}
	\caption{Visualization of SMG-Attention.}
	\label{Fig:1}       
\end{figure}

\subsubsection{Unseen defect classification guided by semi-supervised k-means}
In IC manufacturing, the introduction of new process steps or layouts often brings previously unseen defects. However, most NCD methods rely on two impractical assumptions: the unlabeled set contains no samples from known classes, and the number of unknown classes is given a priori. These assumptions limit applicability in dynamic production lines. To overcome this, we introduce a semi-supervised k-means–based estimator during training that automatically infers the total number of unknown defect classes in the unlabeled data.

We consider a partially labeled dataset $D = D_L \cup D_U$. 
The labeled set $D_L=\{(x_i,y_i)\}_{i=1}^{N_L}$ contains samples only from the base classes $C_{\text{base}}$. The unlabeled set $D_U=\{x_j\}_{j=1}^{N_U}$ may contain samples from both $C_{\text{base}}$ and the unseen classes $C_{\text{unseen}}$.

Our goal is to estimate the total number of classes $\hat{K}$ and thereby obtain the number of unknown classes $\hat{C}_u = \hat{K} - C_l$, which sets the correct output dimension for the downstream classifier. All samples are embedded using the feature extractor $f(\cdot)$ trained with the teacher–student framework and SMG-Attention:

\[
z = f(x) \in \mathbb{R}^d, \quad 
Z_L = \{z_i\}, \; Z_U = \{z_j\}, \; Z = Z_L \cup Z_U.
\]  

For a candidate number of clusters $k$, we run semi-supervised k-means on the full set $Z$. We initialize the $C_l$ centroids using the class means from $D_L$ and anchor them during iterations so that labeled samples remain assigned to their ground-truth centroids. The remaining $k - C_l$ centroids are initialized on $Z$ using k-means++. Unlabeled samples are assigned to the nearest centroid, and centroids are updated until convergence. After convergence, we record the cluster labels $\hat{C}(k)$ obtained on the labeled set $D_L$.

To evaluate clustering quality, we compute accuracy only on the labeled subset $D_L$. We use the Hungarian algorithm to optimally match predicted clusters to ground-truth labels, denoted by a mapping $m$, and compute accuracy:

\[
ACC(k) = \max_{m} \frac{1}{|D_L|} \sum_{(x_i, y_i) \in D_L} 
\mathbf{1}\big\{ y_i = m(\hat{C}_i(k)) \big\}.
\]  

We then search over a predefined range SSS for the $k$ that maximizes $ACC(k)$:

\[
\hat{K} = \arg \max_{k \in S} ACC(k), 
\qquad \hat{C}_u = \hat{K} - C_l.
\]  

For efficiency, we adopt Brent’s method to maximize $ACC(k)$ within $S$, and take the optimal $\hat{K}$ as the final estimate of the total number of classes.

After obtaining $\hat{K}$, we construct classification heads for self-supervised training. For each defect-centered crop, we generate two augmented views and feed them into a shared encoder $E$ along with two independent heads. The encoder is built on the previously trained SMG-Attention and remains frozen in this stage. The two heads are a labeled head $h$ for known classes and an unlabeled head $g$ for unknown classes. We optimize only the classification layers, and both heads together with their projection share parameters across views. The output dimension of the heads is:

\[
C \;=\; C_{\text{labeled}} \;+\; C_{\text{unlabeled}} \;+\; 1,
\]

where the final term corresponds to the defect-free class.

For labeled samples in $D_L$, the labeled head $h$ outputs logits that are supervised with cross-entropy against the one-hot ground-truth labels. For unlabeled samples in $D_U$, the unlabeled head $g$ produces logits that are assigned online pseudo-labels using the Sinkhorn–Knopp algorithm, followed by cross-entropy training.

Within self-supervised training, to encourage unlabeled samples to form novel clusters rather than being absorbed by known classes, we force the teacher’s logits on known classes to zero when processing unlabeled inputs. To stabilize the learning signal, we also use different softmax temperatures for teacher and student: the teacher adopts a lower, sharpening temperature to produce higher-confidence targets, while the student uses a higher, smoothing temperature to better match the teacher’s outputs. This improves cross-view consistency and enhances discrimination of unknown defect classes.

\section{Experiment}
\subsection{Dataset}
The IC-SEM dataset comprises 2,990 images collected from three key manufacturing stages: back end of line (BEOL), deposition (DEP), and dummy poly remove (DPR). It spans 15 defect categories and reflects highly structured backgrounds with localized anomalies under realistic conditions, providing comprehensive support for process-quality evaluation.

\textbf{BEOL} forms metal interconnects and dielectric structures for signal routing and device-level reliability. This subset includes 1,290 images across 6 defect types and is used to assess stability and consistency during interconnect formation.

\textbf{DEP} deposits dielectric or metal films that underpin subsequent etching and patterning and directly affect electrical characteristics. It contains 2 defect types with 775 images and supports analyses of film uniformity and quality control.

\textbf{DPR} removes temporary dummy polysilicon to prepare surfaces for metallization or packaging while preserving layout fidelity. The subset covers 7 defect types with 925 images and facilitates evaluation of removal efficiency and surface cleanliness.

\subsection{Experimental Setup}

\textbf{Backbone and Hyperparameters.} We use ViT-Base/14 pretrained with DINOv3 as the backbone. The normal-information extractor aggregates features with 6 learnable normal tokens, and all inputs are resized to $448\times448$. Hyperparameters are selected via grid search with $\gamma=3.0$ and $\lambda=20$. Training is performed on four NVIDIA RTX 4090 GPUs using AdamW with a learning rate of $1\mathrm{e}{-3}$ for 300 epochs.

\textbf{Evaluation Metrics for Detection.} We report image-level AUROC(i-AUROC) for anomaly recognition and pixel-level AUROC(p-AUROC) for localization accuracy.

\textbf{Evaluation Metrics for Classification.} For classification, we use F1, normalized mutual information (NMI), and adjusted Rand index (ARI). To avoid label-permutation bias, predictions are optimally matched to ground truth with the Hungarian algorithm.

\textbf{Generalization Evaluation Protocol.} To assess generalization, we conduct cross-process evaluation. For detection, we train on BEOL and test on DEP/DPR, and conversely train on DPR and test on BEOL. For classification, we simulate mixed known/unknown settings by treating BEOL as known and DEP/DPR as unknown, and also the reverse where BEOL is unknown and DPR is known. This protocol targets robustness to new processes and layouts.

\subsection{Comparative Experiments}
We compare our method against state-of-the-art few-shot and zero-shot detectors, as shown in Table 1. Our approach outperforms both memory-bank-based methods and reconstruction-based methods. The gain stems from deriving normal cues directly from each test image rather than external support sets, which yields context consistent positives and mitigates cross process feature misalignment.

\begin{table*}[!htb]
\caption{Image-level (i-AUROC) and pixel-level (p-AUROC) on three IC process stages (\%).}
\centering
\tiny
\setlength{\tabcolsep}{6pt}
\begin{tabular}{l c c c c c c}
\noalign{\hrule height 1.2pt}
Method & \multicolumn{2}{c}{BEOL} & \multicolumn{2}{c}{DEP} & \multicolumn{2}{c}{DPR} \\
      & i-AUROC & p-AUROC & i-AUROC & p-AUROC & i-AUROC & p-AUROC \\
\hline
AA-CLIP\cite{ma2025aa} & 95.69 & 93.12 & 97.85 & 96.04 & 73.08 & 89.42 \\
winCLIP\cite{jeong2023winclip}      & 65.08 & 85.62 & 90.86 & 93.25 & 64.69 & 88.24 \\
MUSC\cite{li2024musc}         & 92.24 & 95.04 & 98.01 & 97.02 & 82.77 & 94.12 \\
AnomalyCLIP\cite{zhou2023anomalyclip}  & 82.10 & 95.40 & 90.90 & 91.10 & 73.70 & 92.16 \\
MAE-IC\cite{lu2023masked}       & 81.70 & 90.20 & 97.60 & 95.20 & 82.60 & 91.20 \\
\textbf{Ours} & \textbf{97.58} & \textbf{96.37} & \textbf{99.19} & \textbf{98.92} & \textbf{97.62} & \textbf{96.62} \\
\noalign{\hrule height 1.2pt}
\end{tabular}
\end{table*}

To further validate localization quality, we provide visual comparisons. As shown in Fig. 9, self-normal-information–guided detection focuses more precisely on true defect regions and markedly suppresses background-induced false positives, confirming its effectiveness and robustness on IC SEM
 images.

\begin{figure}[ht]
        \centering
	\includegraphics[width=0.8\textwidth]{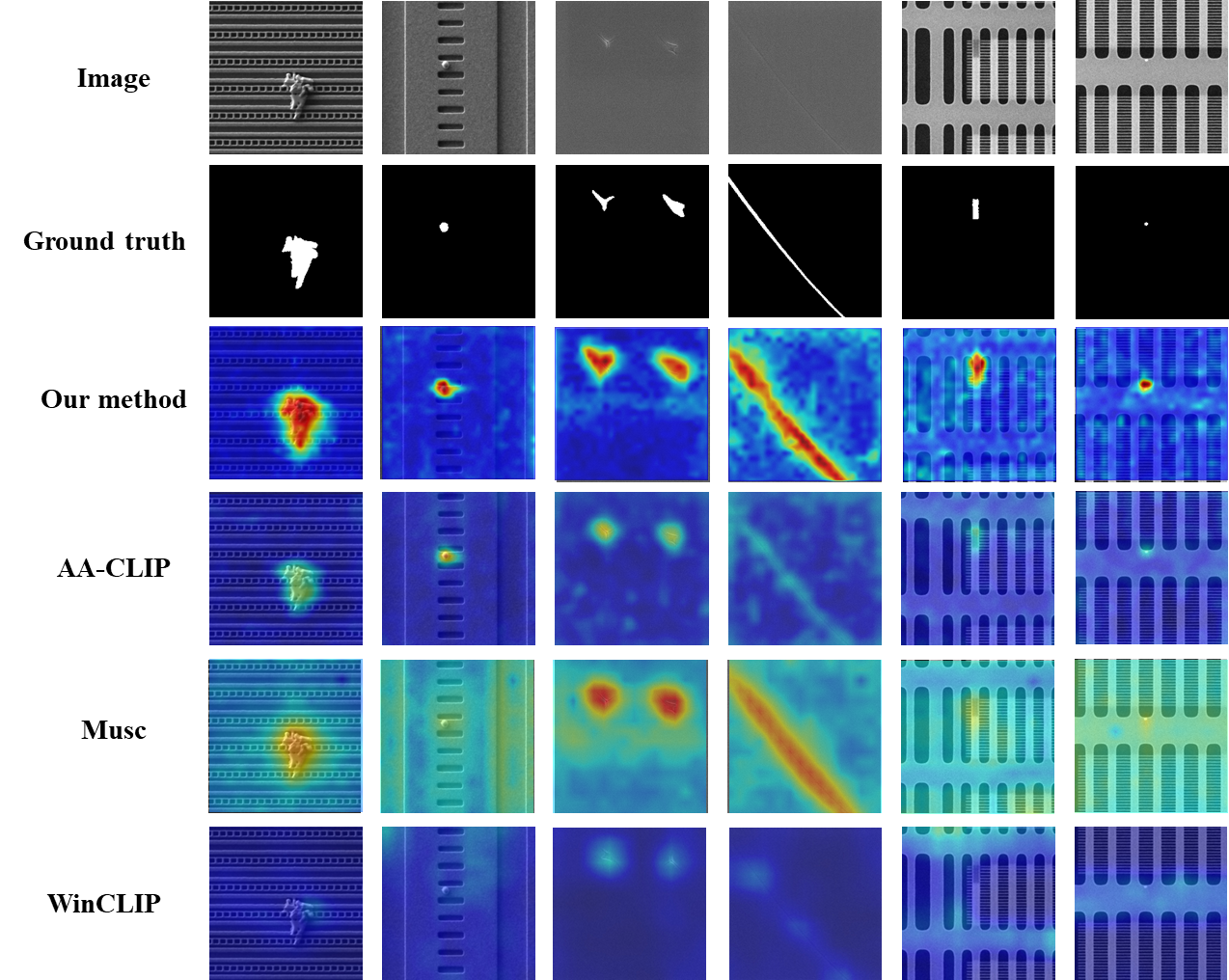}
	\caption{Visualization of different IC defect detection method.}
	\label{Fig:3}       
\end{figure}

For defect classification, after estimating the class number $k$, we compare with multiple classifiers. As shown in Table 2, the classifier trained with SMG-Attention and self-supervision consistently surpasses unsupervised baselines on all metrics and even exceeds some fully supervised baselines. This indicates that soft-mask–guided attention strengthens defect-centric feature learning and improves discrimination of unseen defect categories.

\begin{table*}[htbp]
\tiny
\centering
\setlength\tabcolsep{3pt}
\caption{Performance comparison across BEOL, DEP, and DPR datasets.}
\begin{tabular}{l ccc ccc ccc}
\hline
Method & \multicolumn{3}{c}{BEOL} & \multicolumn{3}{c}{DEP} & \multicolumn{3}{c}{DPR} \\
 & NMI & ARI & F1 & NMI & ARI & F1 & NMI & ARI & F1 \\
\hline
ViT\cite{dosovitskiy2020image} & 0.6637 & 0.6041 & 0.7034 & 0.8084 & 0.7981 & 0.8856 & 0.8813 & 0.9132 & 0.8924 \\
swin-transformer\cite{liu2021swin} & 0.6817 & 0.6221 & 0.7234 & 0.8174 & 0.8071 & 0.8956 & 0.8903 & 0.9222 & 0.9024 \\
IC-DETR\cite{zhao2024integrated} & 0.6907 & 0.7311 & 0.7334 & 0.8219 & 0.8116 & 0.9006 & 0.8948 & 0.9357 & 0.8974 \\
simGCD\cite{wen2023SimGCD} & 0.4520 & 0.3460 & 0.5690 & 0.6688 & 0.6757 & 0.7042 & 0.7221 & 0.7758 & 0.8217 \\
GCD\cite{vaze2022GCD} & 0.5507 & 0.6529 & 0.4195 & 0.8148 & 0.8233 & 0.8898 & 0.8502 & 0.8914 & 0.8766 \\
promptCAL\cite{zhang2023promptcal} & 0.6362 & 0.6825 & 0.6717 & 0.7531 & 0.7368 & 0.9221 & 0.7891 & 0.8294 & 0.8152 \\
UNO\cite{fini2021UNO} & 0.5866 & 0.7542 & 0.5133 & 0.6574 & 0.6793 & 0.7022 & 0.8476 & 0.8821 & 0.8521 \\
AMEND\cite{banerjee2024amend} & 0.6291 & 0.6749 & 0.6642 & 0.6977 & 0.7285 & 0.9087 & 0.6084 & 0.7562 & 0.7092 \\
DCCL\cite{pu2023DCCL} & 0.6102 & 0.6714 & 0.6402 & 0.7825 & 0.8889 & 0.9036 & 0.6036 & 0.6861 & 0.7251 \\
AnomalyNCD\cite{huang2025anomalyncd} & 0.6521 & 0.7733 & 0.6915 & 0.8092 & 0.8911 & 0.9207 & 0.7676 & 0.8031 & 0.7849 \\
MultiADS\cite{sadikaj2025multiads} & 0.6815 & 0.7171 & 0.7097 & 0.7963 & 0.9021 & 0.9147 & 0.8413 & 0.8873 & 0.8642 \\
\textbf{Ours} & \textbf{0.7026} & \textbf{0.8236} & \textbf{0.7462} & \textbf{0.8477} & \textbf{0.9232} & \textbf{0.9660} & \textbf{0.9109} & \textbf{0.9632} & \textbf{0.9341} \\
\hline
\end{tabular}
\end{table*}

\subsection{Ablation Studies}
We conduct systematic ablations to verify the contribution of each core component under a unified train/test split for fair comparison. First, we assess detection with scarce data. As shown in Table 3, even with very few training samples, our method maintains stable segmentation performance, showing that self-normal information extraction reduces reliance on labeled data and well-suited production settings with limited annotations.

\begin{table*}[!ht]
\caption{Effect of training sample size (k-shot) on i-AUROC and p-AUROC across three IC process stages.}
\centering
\tiny
\setlength\tabcolsep{3pt}
\begin{tabular}{l c c c c c c}
\noalign{\hrule height 1.2pt}
 & \multicolumn{2}{c}{BEOL} & \multicolumn{2}{c}{DEP} & \multicolumn{2}{c}{DPR} \\
 & i-AUROC & p-AUROC & i-AUROC & p-AUROC & i-AUROC & p-AUROC \\
\hline
k=1      & 92.69 & 90.11 & 96.71 & 94.62 & 82.16 & 79.98 \\
k=4      & 95.32 & 92.97 & 97.64 & 96.92 & 92.57 & 89.71 \\
full-shot& 97.58 & 96.37 & 99.19 & 98.92 & 97.62 & 94.51 \\
\noalign{\hrule height 1.2pt}
\end{tabular}
\end{table*}

We then analyze the effect of the number of normal tokens. As shown in Fig. 10, performance improves and saturates as the token count increases, achieving the best trade-off at 6. This suggests a moderate number of tokens sufficiently covers diverse normal patterns, while too few limits capacity and too many adds redundancy and higher computational cost.

\begin{figure}[ht]
        \centering
	\includegraphics[width=0.6\textwidth]{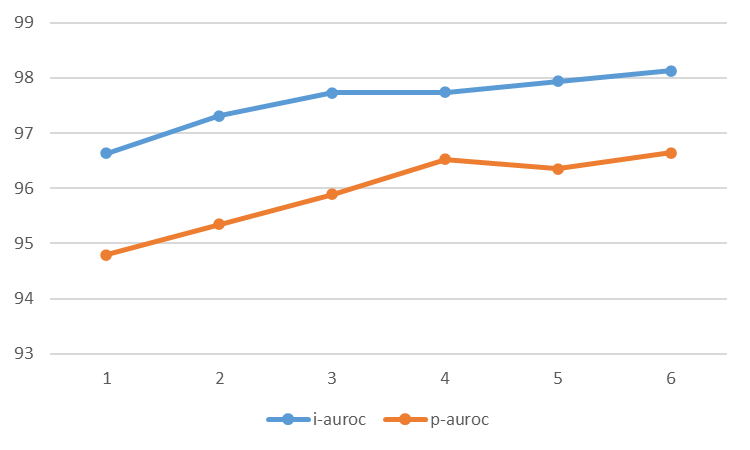}
	\caption{Effect of NI Counts.}
	\label{Fig:5}       
\end{figure}

To validate the reconstruction strategy, we compare it with direct feature matching. Results in Table 4 and Fig. 11 show that direct matching is more sensitive to background noise, whereas reconstruction-based detection is more stable under complex structures, indicating that learning normal patterns via reconstruction better captures subtle normal–abnormal deviations.

\begin{table*}[!ht]
\caption{Comparison between direct Normal-Information (NI) matching and NI-guided reconstruction on three IC process stages (\%). Metrics are image-level (i-AUROC) and pixel-level (p-AUROC).}
\centering
\tiny
\setlength\tabcolsep{3pt}
\begin{tabular}{l c c c c c c}
\noalign{\hrule height 1.2pt}
Method & \multicolumn{2}{c}{BEOL} & \multicolumn{2}{c}{DEP} & \multicolumn{2}{c}{DPR} \\
      & i-AUROC & p-AUROC & i-AUROC & p-AUROC & i-AUROC & p-AUROC \\
\hline
Direct NI matching & 95.82 & 95.41 & 98.79 & 98.12 & 94.67 & 92.51 \\
\textbf{Ours} & \textbf{97.58} & \textbf{96.37} & \textbf{99.19} & \textbf{98.92} & \textbf{97.62} & \textbf{94.62} \\
\noalign{\hrule height 1.2pt}
\end{tabular}
\end{table*}

\begin{figure}[ht]
        \centering
	\includegraphics[width=0.8\textwidth]{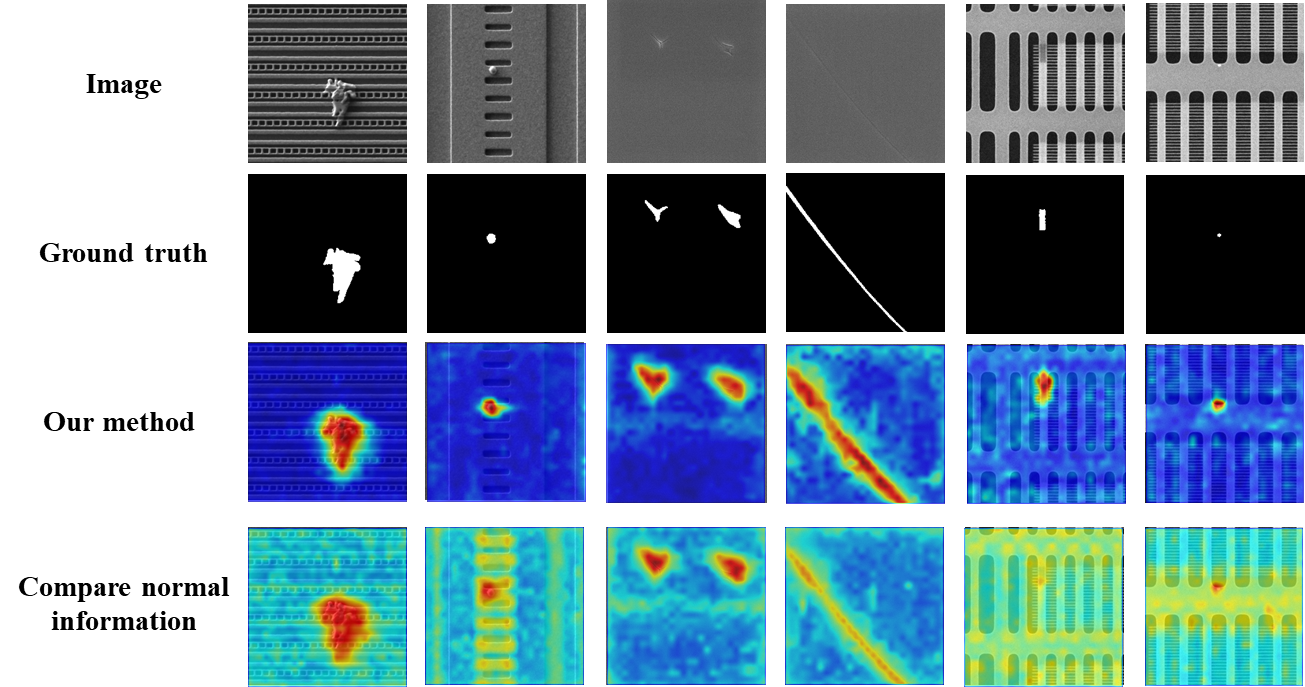}
	\caption{Visual comparison between our normal-information (NI)-guided reconstruction method and a baseline that segments defects by directly comparing NI with local features.}
	\label{Fig:4}       
\end{figure}

\begin{table}[!ht]
  \centering
  \tiny
  \caption{Estimation of the number of classes in unlabelled data.}
  \label{tab:results}
  \setlength{\tabcolsep}{8pt} 
  \renewcommand{\arraystretch}{1.1} 
  \begin{tabular}{lccc}
  \noalign{\hrule height 1.2pt}
    & BEOL & DEP & DPR \\
    \hline
    Ground truth & 8 & 3 & 8 \\
    Ours         &  8 & 3 & 9 \\
    Error        & 0\% & 0\% & 12.5\% \\
  \noalign{\hrule height 1.2pt}
  \end{tabular}
\end{table}

We further evaluate the effectiveness of semi-supervised k-means for estimating the number of unknown defect classes. Table 5 reports estimation errors across the three process datasets. Even when known and unknown samples are mixed, the method accurately infers the true class count with low mean absolute error. This shows that semi-supervised k-means effectively leverages constraints from limited labels while discovering latent structure in unlabeled data, overcoming the need to pre-specify class counts and providing a reliable basis for downstream classification.

\begin{table}[!htbp]
\centering
\tiny
\caption{Comparison of different binarization methods on BEOL.}
\begin{tabular}{lccc}
\hline
\textbf{Method} & \textbf{NMI} & \textbf{ARI} & \textbf{F1} \\
\hline
threshold=0.5 & 0.5412 & 0.6941 & 0.5978 \\
Otsu's method & 0.6297 & 0.7412 & 0.6509 \\
\textbf{ours} & \textbf{0.7026} & \textbf{0.8236} & \textbf{0.7462} \\
\hline
\end{tabular}
\end{table}

To address saliency disparity across defects, we compare adaptive binarization with fixed thresholds. Table 6 shows clear gains in classification, demonstrating that our method dynamically adjusts the segmentation threshold to provide more accurate defect masks for classification.

\begin{table}[htbp]
\centering
\tiny
\caption{Ablation study of masking and cropping on BEOL.}
\begin{tabular}{lccc}
\hline
\textbf{Method} & \textbf{NMI} & \textbf{ARI} & \textbf{F1} \\
\hline
without mask and crop & 0.6637 & 0.6041 & 0.7034 \\
without mask & 0.6257 & 0.7271 & 0.6423 \\
binarized mask & 0.6401 & 0.7937 & 0.6903 \\
\textbf{SMG-ViT} & \textbf{0.7026} & \textbf{0.8236} & \textbf{0.7462} \\
\hline
\end{tabular}
\end{table}

We conduct an ablation on SMG-ViT to isolate the contribution of each component and to enable a fair comparison with a vanilla ViT. “without mask and crop” corresponds to training a vanilla ViT on full images with neither defect-centered cropping nor any spatial prior. “without mask” trains the same ViT on defect-centered crops but still omits any spatial prior, isolating the effect of cropping. “binarized mask” keeps cropping and injects a binary mask vector as the spatial prior in place of the soft mask vector used by SMG-ViT. SMG-ViT combines defect-centered cropping with soft-mask–guided attention. The results show that SMG-ViT preserves contextual cues while better focusing learning on defect regions, achieving the best NMI, ARI, and F1.

For deployment feasibility, our method achieves an average inference time of 28.12 ms per image on a single NVIDIA RTX 4090, meeting real-time requirements on production lines and highlighting practical value.

\section{Industrial Deployment}
We evaluate IC-DefectNCD using SEM review images collected on KLA SEM G* series tools and conduct IC defect detection experiments on these datasets. The software was independently developed by our team and has been delivered to Amedac as a deployable application. As shown in Fig. 12, in Amedac’s production workflow the module ingests SEM images, generates residual heatmaps and adaptive masks on a local workstation GPU, and returns defect classes through a lightweight interface to assist yield engineers in rapid disposition.

\begin{figure}[ht]
        \centering
	\includegraphics[width=0.6\textwidth]{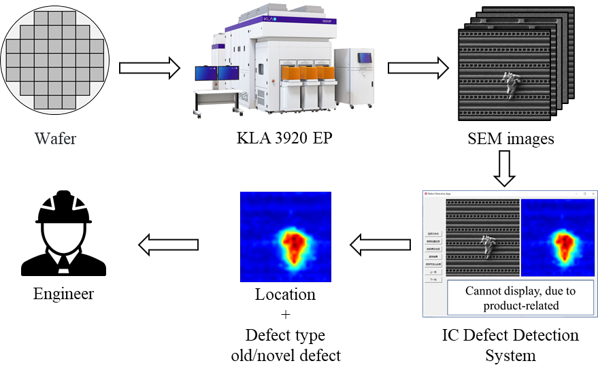}
	\caption{Visual comparison between our normal-information (NI) guided reconstruction method and the baseline that segments defects by directly comparing NI with local features.}
	\label{Fig:4}       
\end{figure}

To assess cross-vendor generalization, we conduct pilot tests with MEGAROBO on SEM images acquired from non-KLA platforms. In our deployment, per-image latency on a single workstation GPU satisfies real-time review constraints, and the interface presents confidence scores, masks, and class names for rapid disposition. The pipeline preserves data locality, supports audit logging and versioned recipes, and can be extended to additional KLA inspection and review modules as needed. This behavior reflects the method’s robustness to layout and process shifts and its ability to adapt to images captured on different tools, underscoring its deployment value and practical generalization.

\section{conclusion}
In this work, we present IC-DefectNCD, a unified framework for defect detection and novel class discovery in IC SEM images, designed to tackle key challenges in semiconductor manufacturing. Our core insight is to leverage single-image internal information, eliminating reliance on external support sets used in prior methods. The framework is built on mining intra-image cues. Self-Normal Information Guided IC Defect Detection performs normal-information guided reconstruction with residual-based localization, and Self-Saliency-driven Adaptive Binarization reliably extracts defect-centered crops. For classification, the Self-Defect Information Guided IC Defect Classification employs the SMG-ViT model to enhance defect-centered feature representations while preserving critical contextual. We introduce a teacher-student model coupled with semi-supervised k-means to automatically discover and classify unseen categories without prior of the class count. Comprehensive experiments on datasets covering BEOL, DEP, and DPR, demonstrate strong performance in both segmentation and classification.

\bibliographystyle{plain}
\bibliography{sample}

\end{document}